\documentclass[fleqn,10pt]{wlscirep}
\usepackage{tabularx}
\usepackage[utf8]{inputenc}
\usepackage{multirow} 
\usepackage[T1]{fontenc}
\usepackage{lineno}
\usepackage{hyperref}
\hypersetup{colorlinks,allcolors=black}
\usepackage{array} 
\newcolumntype{Y}[1]{>{\raggedleft\arraybackslash}m{#1}}

\title{DeepExtremeCubes: Integrating Earth system spatio-temporal data for impact assessment of climate extremes}

\author[1,2, *]{Chaonan Ji}
\author[3]{Tonio Fincke}
\author[4,5]{Vitus Benson}
\author[6]{Gustau Camps-Valls}
\author[6]{Miguel-Ángel Fernández-Torres}
\author[4]{Fabian Gans}
\author[1,2]{Guido Kraemer}
\author[1,2,7]{Francesco Martinuzzi}
\author[1,2,8]{David Montero}
\author[1,2]{Karin Mora}
\author[6]{Oscar J. Pellicer-Valero}
\author[4,5]{Claire Robin}
\author[1,2,9]{Maximilian Söchting}
\author[4]{Mélanie Weynants}
\author[1,2,8,10,*]{Miguel D. Mahecha}

\affil[1]{ Remote Sensing Centre for Earth System Research (RSC4Earth), Leipzig University, Leipzig, 04103, Germany}
\affil[2]{ Institute for Earth System Science and Remote Sensing, Leipzig University, Leipzig, 04103, Germany}
\affil[3]{Brockmann Consult GmbH, Hamburg, 21029, Germany}
\affil[4]{Max Planck Institute for Biogeochemistry, Jena, 07745, Germany}
\affil[5]{ELLIS Unit Jena, Jena, 07745, Germany}
\affil[6]{Image Processing Laboratory (IPL), Universitat de Val\`encia, Val\`encia, 46980, Spain}
\affil[7]{Center for Scalable Data Analytics and Artificial Intelligence (ScaDS.AI), Leipzig, 04105, Germany}
\affil[8]{German Centre for Integrative Biodiversity Research (iDiv) Halle-Jena-Leipzig, Leipzig, 04103, Germany}
\affil[9]{Image and Signal Processing Group, Leipzig University, Leipzig, 04109, Germany}
\affil[10]{Helmholtz Centre for Environmental Research (UFZ), Leipzig, 04318, Germany}

\begin{abstract}

With climate extremes' rising frequency and intensity, robust analytical tools are crucial to predict their impacts on terrestrial ecosystems. Machine learning techniques show promise but require well-structured, high-quality, and curated analysis-ready datasets. Earth observation datasets comprehensively monitor ecosystem dynamics and responses to climatic extremes, yet the data complexity can challenge the effectiveness of machine learning models. Despite recent progress in deep learning to ecosystem monitoring, there is a need for datasets specifically designed to analyse compound heatwave and drought extreme impact. Here, we introduce the \emph{DeepExtremeCubes} database, tailored to map around these extremes, focusing on persistent natural vegetation. It comprises over 40,000 spatially sampled small data cubes (i.e. minicubes) globally, with a spatial coverage of 2.5 by 2.5 km. Each minicube includes (i) Sentinel-2 L2A images, (ii) ERA5-Land variables and generated extreme event cube covering 2016 to 2022, and (iii) ancillary land cover and topography maps. The paper aims to (1) streamline data accessibility, structuring, pre-processing, and enhance scientific reproducibility, and (2) facilitate biosphere dynamics forecasting in response to compound extremes.

\end{abstract}
\begin{document}
\flushbottom
\maketitle
\thispagestyle{empty}

\section*{Background \& Summary}
There has been an unprecedented rise in the frequency and severity of climate extremes\cite{fischer2023storylines}. These rising extremes can have severe ecological\cite{flach2021vegetation} and socio-economic consequences\cite{lee2024reclassifying}, challenging our established paradigms of climate science\cite{linkov2014changing}. For instance, in 2018, central and northern Europe experienced a record-breaking Compound Heatwave and Drought (CHD) event, which extensively impacted agriculture, forests, water supply, and the socio-economic sector\cite{rousi2023extremely}.
Given the increasing intensity and adverse impacts of CHD events in the warming climate, it is critical to understand their intricate dynamics and interactions with climate drivers, spatial conditions, timing, and terrestrial ecosystems\cite{olesen2002consequences, seneviratne2012changes, zscheischler2018future, mahecha2023biodiversity}.

The exponential increase in Earth observation data represents a significant advancement but also introduces complex data management and analysis challenges\cite{mahecha2020earth, montero2023data}. In an era marked by rapid advances in remote sensing capabilities, including satellite observations, aerial imaging, and ground-based records, researchers have access to unprecedented amounts of information. These data are crucial for understanding the impacts of climate extremes\cite{yang2013role, smith2014remote, mahecha2017detecting, massaro2023spatially}. Effective sampling strategies are required to harness this data deluge, ensuring relevance and manageability. Data cubes provide a flexible and efficient way to organise and analyse large volumes of multidimensional data, making such datasets manageable and streamlined across variables and spatio-temporal scales\cite{mahecha2020earth,montero2024ondemand}.


Machine Learning (ML) has been introduced into climate science as a valuable tool to understand and predict climate extremes and their impacts, as well as to decipher the interactions between climate and ecosystems \cite{reichstein2019deep, CampsValls21wiley, martinuzzi2023learning, montero2024recurrent_gpp, beigaite2022identifying, aidetoolbox}. Moreover, Deep Learning (DL) allows the identification of complex patterns and correlations that might elude traditional data science methods, thereby helping scientists to better understand the underlying mechanisms of climate variability and change. However, since ML generally performs best with large sample sizes, extreme impact prediction often has significantly smaller sample sizes compared to non-extreme conditions, which complicates the application of ML. In this dataset, we tackle this issue by oversampling extreme areas using the minicube strategy, which has a large distribution in space instead of time. This method introduces additional biases, as ML tends to amplify them. Nonetheless, this trade-off is necessary and must be considered when training models.

The sophisticated Earth observation databases that train ML models for analysing climate extremes are growing. These datasets primarily focus on addressing the scarcity of curated data concerning complex weather patterns and climate extremes' impacts on ecosystems. For instance, The ExtremeWeather dataset\cite{racah2017extremeweather} provides labelled extreme weather events (i.e., tropical depression, tropical cyclone, extratropical cyclone, atmospheric river) as boxes, along with climatic and meteorological variables on a global grid of 768 by 1152. This dataset allows training ML models to leverage spatial and temporal information to predict the localisation of extreme weather events. ClimateNet\cite{kashinath2021climatenet} provides an expert-labelled dataset that enables pixel-level identification of extreme events using ML models. Additionally, cross-domain and high-resolution datasets are designed to include localised variables critical for analysing responses to climate extremes, incorporating data from diverse domains. For example, EarthNet2021\cite{requena2021earthnet2021} aims to bridge the data gap by integrating a variety of data variables such as precipitation, temperature, sea-level pressure, digital elevation models, and Sentinel-2 Multi-Spectral Instrument (MSI) images, offering a holistic view of Earth system. A model trained on EarthNet2021 can forecast optical satellite images of high perceptual quality. The newly enhanced version, GreenEarthNet\cite{benson2024multimodal}, focuses more on predicting vegetation dynamics and includes an improved high-quality cloud mask\cite{aybar2022cloudsen12}. The FluxnetEO data cubes\cite{walther2022fluxneteo} provide fully gap-filled Nadir BRDF Adjusted Reflectance (NBAR) data from MODIS, as well as Land Surface Temperature (LST) and several vegetation indices for the Fluxnet sites\cite{baldocchi2019fluxes}, aiming for modelling carbon and water fluxes. Moreover, DynamicEarthNet\cite{toker2022dynamicearthnet} tracks daily land use and land cover changes across 75 global regions from 2018 to 2019, focusing on detecting land cover changes. BigEarthNet\cite{sumbul2021bigearthnet} is a large-scale benchmark dataset consisting of Sentinel-2 satellite images with multi-label land use and land cover. Presto's Training Dataset\cite{tseng2023lightweight} is a high-resolution dataset that provides detailed data for training ML models to significantly improve the prediction and understanding of climate extremes and their impacts. 

However, these multi-purpose initiative datasets do not focus specifically on the impact of CHD extremes. We need harmonised datasets tailored for spatio-temporal ML methodologies, aiming to train ML methods to forecast and explain the impacts of extreme events such as droughts and heatwaves. Given the importance of CHD extremes and challenges arising from data biases and their repercussions, this paper is poised to propose a solution that encapsulates precision and reproducibility. Here, we present the \emph{DeepExtremeCubes} dataset, a collection of minicubes that use a sampling methodology to focus on capturing the impact of CHD extremes globally. Specifically, we introduce 1) a globally stratified sampling procedure, 2) a reproducible data processing pipeline combining multi-modal data, and 3) a representative global dataset to train ML models on CHD extremes, which is analysis-ready and shared in cloud-native format.

\section*{Methods}

The analysis of CHD extremes necessitates examining a broad range of Earth observation variables across climatic, meteorological, ecological, and topographical dimensions at various spatial and temporal scales \cite{mahecha2020earth}. Sampling these relevant datasets is crucial to focus on CHD impacts and to understand the complex interactions of different drivers, spatial conditions, and timing of these processes. The \emph{DeepExtremeCubes} dataset employs sampled minicubes targeted at regions experiencing extreme CHD events and their surroundings, facilitating a more detailed investigation. From a practical viewpoint, managing the vast, high-dimensional Earth system datasets requires significant computational resources. Segmenting these datasets into smaller, manageable subsets (i.e. minicubes) can enhance machine learning computations efficiency\cite{montero2023data}.

\subsection*{Input data sources}
Two categories of input data sources are used to create \emph{DeepExtremeCubes}. One is the reference dataset used to determine the strata. This encompasses the Dry and hot extreme events database (Dheed dataset, a predefined global dataset of CHD extreme events) \cite{weynants_dheed_2024} and the European Space Agency (ESA) Climate Change Initiative (CCI) land cover map. The other category of data sources consists of the comprehensive Earth system datasets from which the data included in individual minicubes is extracted. We first introduce the reference datasets and then provide details on the comprehensive datasets within the generated minicubes.

\subsubsection*{Dheed event detection dataset}
\begin{figure}[ht]
\centering
\includegraphics[width=0.8\linewidth]{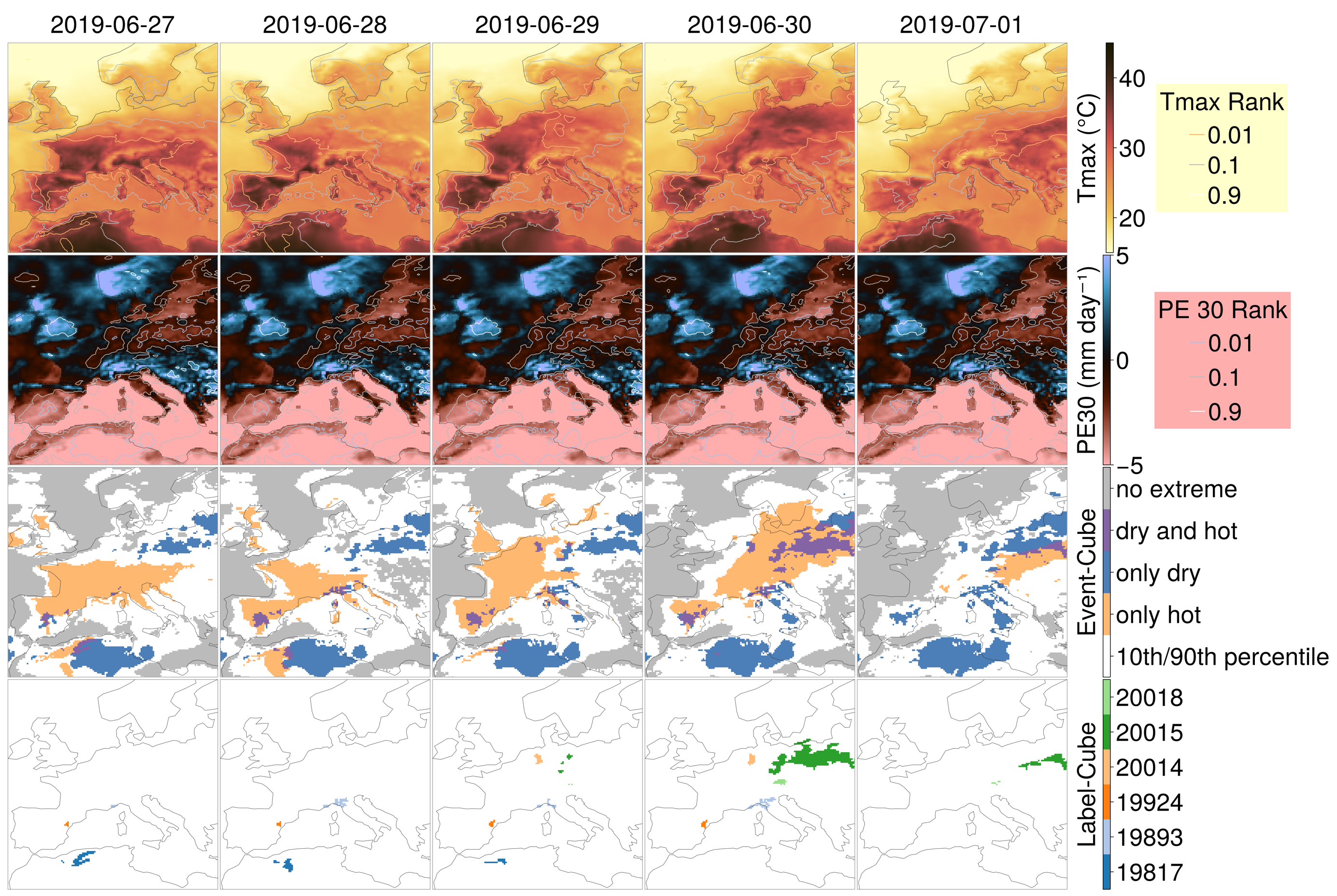}
\caption{An example of CHD detection in the Dheed database showing the evolution across Europe of the maximum daily temperature (Tmax, top row), the Precipitation - Evapotranspiration balance averaged over the previous 30 days (PE30, second row), along with the threshold of 0.01 on the ranked values used to detect extremes (Rank, first and second rows), the synthesis of the four indicators (Event-Cube, third row) and the labelled CHD events lasting at leat three days (Label-Cube, fourth row) from 2019-06-27 to 2019-07-01.}
\label{fig1}
\end{figure}

Dheed \cite{weynants_dheed_2024} is a database of labelled CHD events utilising atmospheric temperature and precipitation from daily aggregated ERA5-Land reanalysis data. A spatially and temporally example piece of a Dheed dataset is shown in Fig.~\ref{fig1}. The maximum daily Temperature at 2~m (Tmax) is used to detect heatwaves, and the daily differences of Precipitation and Evapotranspiration (PE) averaged over 30, 90, and 180 days (PE30, PE90, PE180) are employed to detect droughts \cite{weynants_dheed_paper_2024}. The Dheed's label-cube covers a time range from 2016-01-01 to 2021-12-31, with a spatial resolution of 0.25\textdegree. Groups of spatio-temporal grid cells with extreme values connected across space and/or time are each assigned a unique event label. Dheed's labelled events have been benchmarked against extreme events documented in the literature or the media.

\subsubsection*{Land cover map}
\label{sec:LandCoverMap}
The ESA CCI land cover dataset employs the GlobCover unsupervised classification chain framework\cite{defourny2023observed} to generate global annual land use maps from 1992 to 2020. It uses a combination of multi-year and multi-sensor strategies, incorporating data from various satellites such as ENVISAT-MERIS (2003–2012), AVHRR (1992–1999), SPOT-Vegetation (1999–2013), and PROBA-Vegetation (2013–2020)\cite{geomatics2017land}. The dataset categorises 37 land cover classes according to the United Nations Land Cover Classification System\cite{rajib2023human} and offers the data at a 300-m spatial resolution in GeoTIFF and NetCDF formats. The selection criteria for a reliable land cover map aim to best meet the requirements to analyse vegetation responses to CHD extremes. First, the map must cover most of the study period from 2016 to 2022, ensuring data continuity to reflect land cover changes. Second, data must be readily accessible for direct download and use. Third, a detailed classification of vegetation types is crucial, particularly focusing on persistent vegetation covers such as broad-leaved trees, needle-leaved trees, and grassland. 
While some vegetation classes were merged to simplify the initial sampling process, it was essential to retain the original and finer classifications in the minicubes for subsequent analyses. Various global land cover maps were evaluated\cite{potapov2022global,song2018global,zanaga2022esa}, but none met these criteria as well as the ESA CCI WorldCover map. 
For instance, the Global Land Analysis and Discovery (GLAD) laboratory's Land Cover and Land Use Change (LCLUC) data\cite{potapov2022global}, based on Landsat, offers high accuracy in certain non-forest regions and detailed classifications of open canopy forests in Africa. However, its infrequent updates (every five years) during 2000 and 2020 and limited vegetation classification hinder its suitability as the reference land cover map for the \emph{DeepExtremeCubes} dataset. The ESA CCI WorldCover map\cite{zanaga2022esa} offers high-resolution data for 2020 and 2021. However, it does not cover the entire study period and lacks comprehensive tree-type classifications. Therefore it is not suitable for the detailed vegetation analysis required for this study.

\subsubsection*{Data sources within each minicube}
In addition to subsets of the Dheed dataset and the CCI land cover map, each minicube contains multiple data modalities: (1) Sentinel-2 MSI surface reflectance (L2A) time series data \cite{drusch2012sentinel}, (2) a corresponding deep-learning-based cloud mask \cite{aybar2022cloudsen12, benson2024multimodal}, (3) ERA5-Land meteorological reanalysis variables \cite{munoz2021era5}, and (4) data from the Copernicus Digital Elevation Model (DEM) \cite{CopernicusDEM}. The ERA5-Land reanalysis data provides information on the historical weather conditions, represented by variables such as temperature, humidity, soil moisture, and others. Sentinel-2 MSI satellite images are a proxy observation for vegetation health. We include bands B02, B03, B04, B05, B06, B07, and B8A, which can be used to compute vegetation indices\cite{zeng2022optical_vis,montero2023asi}. Together, the ERA5-Land and Sentinel-2 data allow us to study the impact of CHD extremes on vegetation in the \emph{DeepExtremeCubes} dataset. In Sentinel-2 images, pixels obscured by clouds and cloud shadows can be difficult to distinguish from actual changes in the underlying ecosystem, which may challenge subsequent analyses' accuracy. By incorporating the EarthNet Cloud Cover Mask\cite{benson2024multimodal}, which is based on the CloudSEN12 dataset\cite{aybar2022cloudsen12}, obscured pixels can be filtered out, ensuring that vegetation biological dynamics are based on clear and reliable optical remote sensing data. In addition, the Copernicus DEM data is included as one of the key factors in climate-vegetation interactions. The Copernicus DEM provides topographical data at 30~m, enabling us to consider how elevation influences local climate conditions, subsurface hydrology and vegetation patterns. It is crucial in regions where elevation varies significantly and is also important on a global scale.
For our minicubes, Sentinel-2 and the cloud mask are spatio-temporal arrays, ERA5-Land is included as a single-pixel time series (temporal array), and the DEM is a static image (spatial array).

\subsection*{Approach}

\begin{figure}[ht]
\centering
\includegraphics[width=0.9\linewidth]{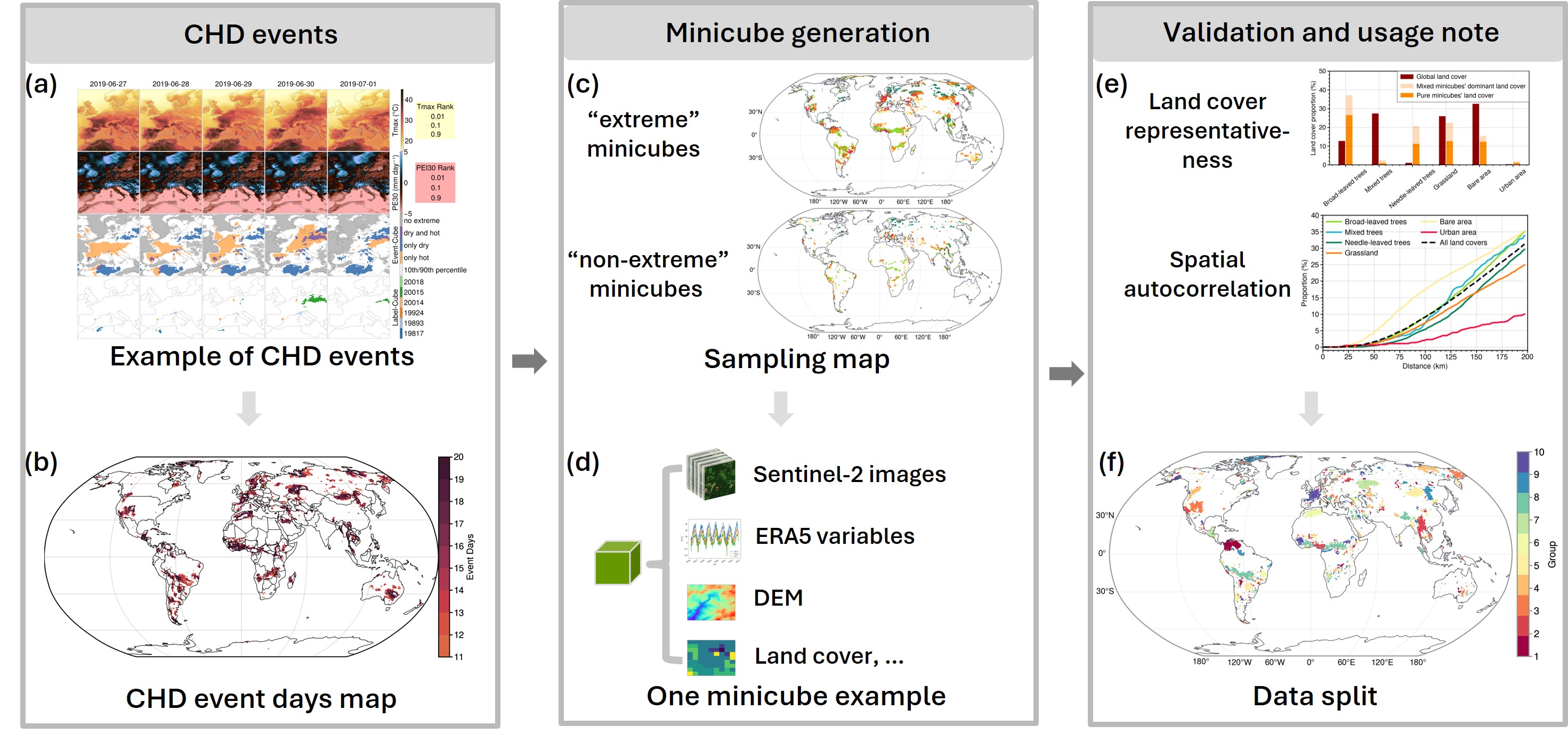}
\caption{A schematic diagram illustrating the \emph{DeepExtremeCubes} database development and validation workflow. It includes: (a) The Dheed dataset presenting an example of a CHD event in this data. (b) The CHD event days map showing locations that experienced 10 or more event days according to the Dheed event detection dataset. (c) Minicubes that experienced 10 or more CHD days ("extreme" minicubes) and those that did not experience any CHD days in the Dheed dataset ("non-extreme" minicubes). (d) A demonstration of a minicube, which includes remote sensing images, climatic and meteorological variables, Digital Elevation Model (DEM), land cover, etc. (e) The representativeness and spatial autocorrelation of the \emph{DeepExtremeCubes} dataset for minicubes with different land cover types. (f) The spatial data split of \emph{DeepExtremeCubes} for potential users to train their forecasting models.}
\label{fig2}
\end{figure}

We incorporated comprehensive data sources to develop the \emph{DeepExtremeCubes} dataset, which comprises minicubes with a spatial size of 2.5 km by 2.5 km, covering the period from 2016 to 2022. The schematic approach is shown in Fig.~\ref{fig2}. The Dheed dataset was used to generate a CHD event days map to determine the sampling locations for the minicubes. Subsequently, the \emph{DeepExtremeCubes} minicubes were created with various variables. Additionally, we prepared a spatial data split strategy for subsequent users to train their ML forecasting models.

\subsubsection*{Sampling locations for minicubes}

The sampling of minicube locations began with identifying areas frequently experiencing CHD extremes. We sampled from both areas affected by extremes and surrounding areas with similar land covers ("extreme" and "non-extreme" locations). This approach allows ML models to learn from both CHD-impacted and non-impacted instances, enhancing prediction accuracy and facilitating accurate estimation of carbon sequestration loss in CHD extremes. Additionally, we adjusted the locations based on land cover, with particular emphasis on persistent vegetation land covers.

The Dheed dataset was compressed to a CHD event days map, marking all pixels that experienced 10 or more CHD event days during 2016 and 2021. Due to the large size of the Dheed dataset, it could not be processed quickly in the following steps. Therefore, we aggregated the temporal dimension and generated the CHD event days map (see Fig.~\ref{fig2})(b). All pixels that experienced 10 or more CHD event days were marked as potential central sampling locations. Around 80\% of the "extreme" minicubes were located in heavily impacted areas (i.e., areas marked in the event days map), while roughly 20\% "non-extreme" minicubes were situated in the vicinity of "extreme" areas and did not experience any CHD events (i.e., areas with 0 event days). This step serves two main purposes: first, it enables the models to learn from CHD-impacted and non-impacted instances, enhancing predictive accuracy. Second, it allows for more accurate computation of carbon sequestration losses within regions covered by minicubes by comparing paired minicubes from both "extreme" and "non-extreme" conditions. To avoid spatial autocorrelation at very close distances in random sampling within "extreme" areas, where sampled locations tend to cluster, we maintained a spatial grid of 0.125° (half of the Dheed dataset's resolution). We selected no more than one sample per grid cell for each land cover type.

We defined a set of target vegetation types that best summarise all the persistent land covers while keeping a focus on vegetation. In this sampling step, we merged most of the land covers to simplify the sampling categories (see details in Table~s1).
These merged vegetation classes include broad-leaved trees, needle-leaved trees, mixed trees, and grassland. To enhance the diversity of land covers and examine prediction accuracy through comparisons between vegetation land covers and other persistent land covers, we also included bare area and urban area. Thereby, we focused on six land cover types in total. In addition, we assessed the purity of land covers within defined minicubes to evaluate the varying behaviours of the ML prediction model concerning pure versus mixed land covers. Given the ESA CCI land cover map's resolution of 300 meters per pixel, a minicube (2.5 km by 2.5 km) encompasses approximately $9\times9$ pixels. We established a spatial window of 81 pixels ($9\times9$), centred on the central pixel, to determine the purity of each minicube's land cover. If 65 or more pixels within this window exhibit the same land cover (equivalent to an $8\times8$ pixel area, about 80\% of the spatial coverage of a minicube), the central pixel is classified under "pure land cover" and is eligible to be the central point of a minicube. Conversely, if fewer than 65 pixels share a single cover type, the central pixel is considered to have "mixed land cover." Given the importance of land cover purity in one minicube for ML prediction models, we set a lower threshold such that the central pixel must display at least 50\% similarity (40.5 pixels in a $9\times9$ matrix). If a land cover meets this threshold range of 50\%-79\%, the central pixel is still considered a potential sampling location and is marked with this land cover as the dominant land cover. We also list the second dominant land cover by tallying the remaining land cover classes and selecting the most frequent. Fig. S1 presents the detailed distribution of (a) the pure land cover map and the mixed land cover maps, including (b) the dominant land cover map and (c) the secondary land cover map.

\begin{figure}[ht]
\centering
\includegraphics[width=\textwidth]{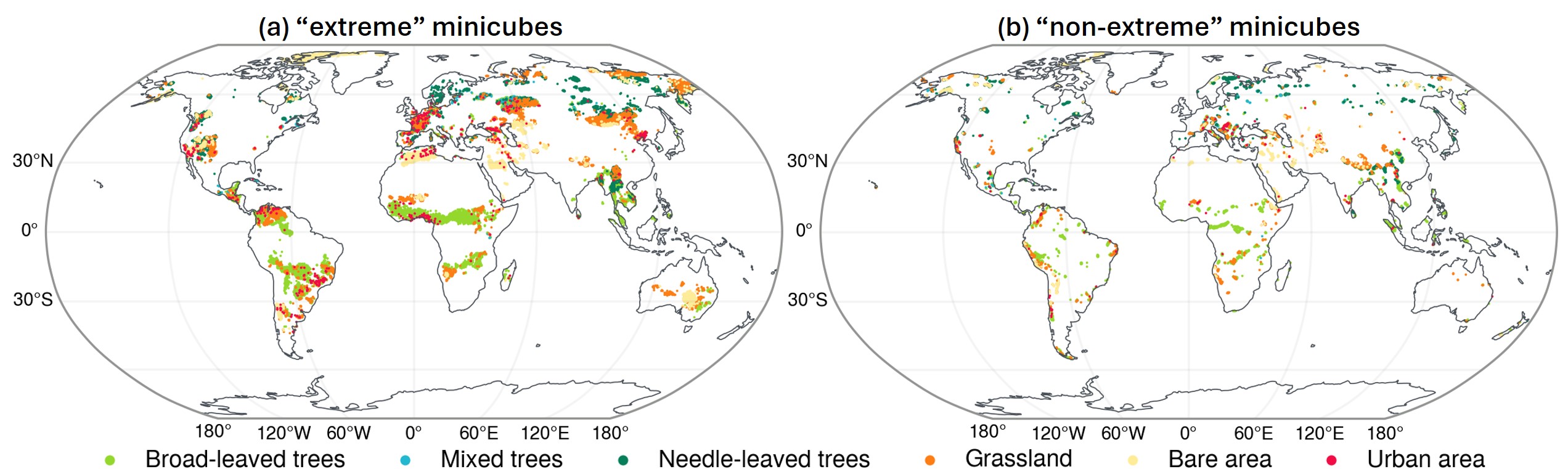}
\caption{The distribution of all sampled minicube locations. (a) Minicubes that experienced 10 or more CHD days ("extreme" minicubes) and (b) minicubes that did not experience CHD events during 2016 and 2021 ("non-extreme" minicubes). The land cover types depicted represent either the pure land cover type for the pure land cover minicubes or the dominant land cover for the minicubes covered by mixed land covers.}
\label{fig3}
\end{figure}

In summary, the minicube location sampling is based on two factors: whether the area is impacted by CHD extremes, and is purely or mixed covered by the target land covers. First, the minicube location sampling primarily focuses on selecting areas affected by CHD extremes, specifically only those with more than 10 event days. From this set 80\% of all mincubes are chosen. The other 20\% is selected from areas surrounding these extremes, which did not experience CHD extremes. We want to distinguish between minicubes covered by pure or mixed land covers, to facilitate exploration of land cover purity for prediction. This categorisation considers pure land cover (about 80\%-100\% covered by one land cover) and mixed land cover (about 50\%-79\% covered by one land cover). When a minicube has mixed coverage, its second land cover class is provided to offer additional contextual information. The results of the minicube sampling are shown in Fig.~\ref{fig3}. The samples indicate the central location of each minicube. 

\subsubsection*{Minicube Generation and variables}

We generated minicubes given the previously established central locations. A minicube is a dataset covering an area of 2.56~km by 2.56~km around the central location, ranging from 2016 to 2022, incorporating elements from various data sources (as listed in Table \ref{table:mc_components_tab}). To minimise the distortion of the original data, we opted to maintain separate spatial and temporal resolutions, which can be harmonised during subsequent processing if necessary. We collected data from various sources during the generation process and documented details about processing steps in configuration files. We consolidated the variables into a single dataset \cite{xcube2023} and stored them in the Zarr format (\hyperlink{https://zarr.readthedocs.io/}{https://zarr.readthedocs.io/}). The generation process comprised two stages: initially, we created a "base-minicube" composed exclusively of Sentinel-2 data. In the second stage, this "base-minicube" was updated with the remaining components. By adopting this strategy, we reduce the frequency of Sentinel-2 data pooling, which is the most resource-intensive part of the process. Additionally, we generate corresponding configuration files for these two phases.

\begin{table}
\centering
\begin{tabular}{
 |>{\centering\arraybackslash}m{0.16\linewidth}|
 >{\centering\arraybackslash}m{0.18\linewidth}|
 >{\raggedleft\arraybackslash}m{0.12\linewidth}|
 >{\raggedleft\arraybackslash}m{0.12\linewidth}|
 >{\raggedleft\arraybackslash}m{0.12\linewidth}|
 >{\raggedleft\arraybackslash}m{0.13\linewidth}|}
\hline
\textbf{Data} & \textbf{Variable} & \textbf{Spatial size/pixel} & \textbf{Spatial resolution/m} & \textbf{Temporal extent} & \textbf{Temporal resolution/day} \\
\hline
CCI land cover map & Land cover class & 9 * 9 & 300 & 2016 & -- -- \\
\hline
Copernicus DEM & DEM & 128 * 128 & 20 & -- -- & -- -- \\
\hline
Sentinel-2 L2A & Bands 2-8A, Scene Classification Layer & 128 * 128 & 20 & 2016.01.01 - 2022.10.10 & 5 \\
\hline
EarthNet cloud mask & Cloud mask & 128 * 128 & 20 & 2016.01.01 - 2022.10.10 & 5 \\
\hline
ERA5-Land data & Min, max, and mean of selected variables & 1 * 1 & -- -- & 2016.01.01 - 2022.10.10 & 5 \\
\hline
Event Data & Event codes and labels & 1 * 1 & -- -- & 2016.01.01 - 2021.12.31 & 1 \\
\hline
\end{tabular}
\caption{The components of a minicube.}
\label{table:mc_components_tab}
\end{table}

\section*{Data Records}

We intend to permanently store the \emph{DeepExtremeCubes} dataset in the \href{https://opensciencedata.esa.int/}{ESA Open Science Catalogue (OSC)} once the platform is operational. 
As the dataset is large, approximately 3.2 TB, preparing the storage in OSC will take a few more months. During this time, we use the Amazon Web Services (AWS) bucket for the manuscript review process, see \href{https://deepextremes-public.s3.eu-central-1.amazonaws.com/readme.pdf}{this access guide} (password: deepextremes). The database includes the \emph{DeepExtremeCubes} minicubes, \href{https://deepextremes-public.s3.eu-central-1.amazonaws.com/mc_registry_v4.csv}{a registry table} providing all attributes within each minicube, and the Dheed dataset. Table~\ref{table2} lists all attributes in the registry table. Users can use it to find minicubes based on specific criteria, such as spatial extent, components, land classes, or labelled extreme events. In addition, we provide \href{https://deepextremes-public.s3.eu-central-1.amazonaws.com/mc_10.22_50.96_1.3_20230928_0.zip}{a minicube example} and \href{https://deepextremes-public.s3.eu-central-1.amazonaws.com/DeepExtremesCubes_Access.ipynb}{a demonstration Jupyter notebook} with direct hyperlinks to help explore the \emph{DeepExtremeCubes} minicubes.

\begin{table}[h]
\centering
\begin{tabular}{|>{\raggedright\arraybackslash}p{3.2cm}|>{\raggedright\arraybackslash}p{13.5cm}|}
\hline
\textbf{Attributes} & \textbf{Description} \\
\hline
mc\_id & Identifier of the minicube within the \emph{DeepExtremeCubes} system with latitude, longitude, generation version, and generation date. \\
\hline
path & Path to the minicube file in the storage system. \\
\hline
location\_source & Source file for the event location data. \\
\hline
location\_id & Geographical identifier associated with the latitude and longitude of the minicube. \\
\hline
version & The generation version of the minicube \\
\hline
type & Type of minicube. Either "full", "backup", or "base", where "full" means that the minicube has all required properties, "backup" refers to cubes that were created not based on the determined events and "base" means they are missing properties. \\
\hline
geometry & Geometric boundary of the minicube. \\
\hline
creation\_date & Date and time when the minicube was created. \\
\hline
modification\_date & Date and time when the minicube was last modified. \\
\hline
events & ID's of detected compound events of heatwaves and droughts were recorded at the minicubes location, along with each respective event's start and end times. \\
\hline
class & The type of land cover in case the minicube occupies $\geq$80\% of one land cover. None if the minicube has mixed land cover. \\
\hline
dominant\_class & The most frequent land cover in case it covers between 50\% and 80\%. None for pure land cover cubes. \\
\hline
second\_dominant\_class & The second most frequent land cover in case of a mixed land cover minicube. None for pure land cover cubes. \\
\hline
s2\_l2\_bands & Version of the Sentinel-2 reflectance data in the minicube. \\
\hline
ERA5-Land & Version of the ERA5-Land climate reanalysis data in the minicube. \\
\hline
cci\_landcover\_map & Version of the CCI land cover data in the minicube. \\
\hline
copernicus\_dem & Version of the Copernicus DEM data in the minicube.\\
\hline
de\_africa\_climatology & Version of climatology data specific to Africa in the minicube (applicable only to some cubes). \\
\hline
event\_arrays & Version of data of recorded events in the minicube. \\
\hline
s2cloudless\_cloudmask & Version of cloud masking data generated using the S2Cloudless model (where applicable). \\
\hline
sen2cor\_cloudmask & Version of cloud masking data generated using the Sen2Cor tool (where applicable). \\
\hline
unetmobv2\_cloudmask & Version of cloud masking data processed using the UnetMobV2 model (where applicable). \\
\hline
remarks & Additional notes. \\
\hline
\end{tabular}
\caption{Summary of minicube attributes in the \emph{DeepExtremeCubes} dataset.}
 \label{table2}
\end{table}

\section*{Technical Validation}

\subsection*{Land cover representation}

\begin{figure}[ht]
 \centering
 \includegraphics[width=0.6\linewidth]{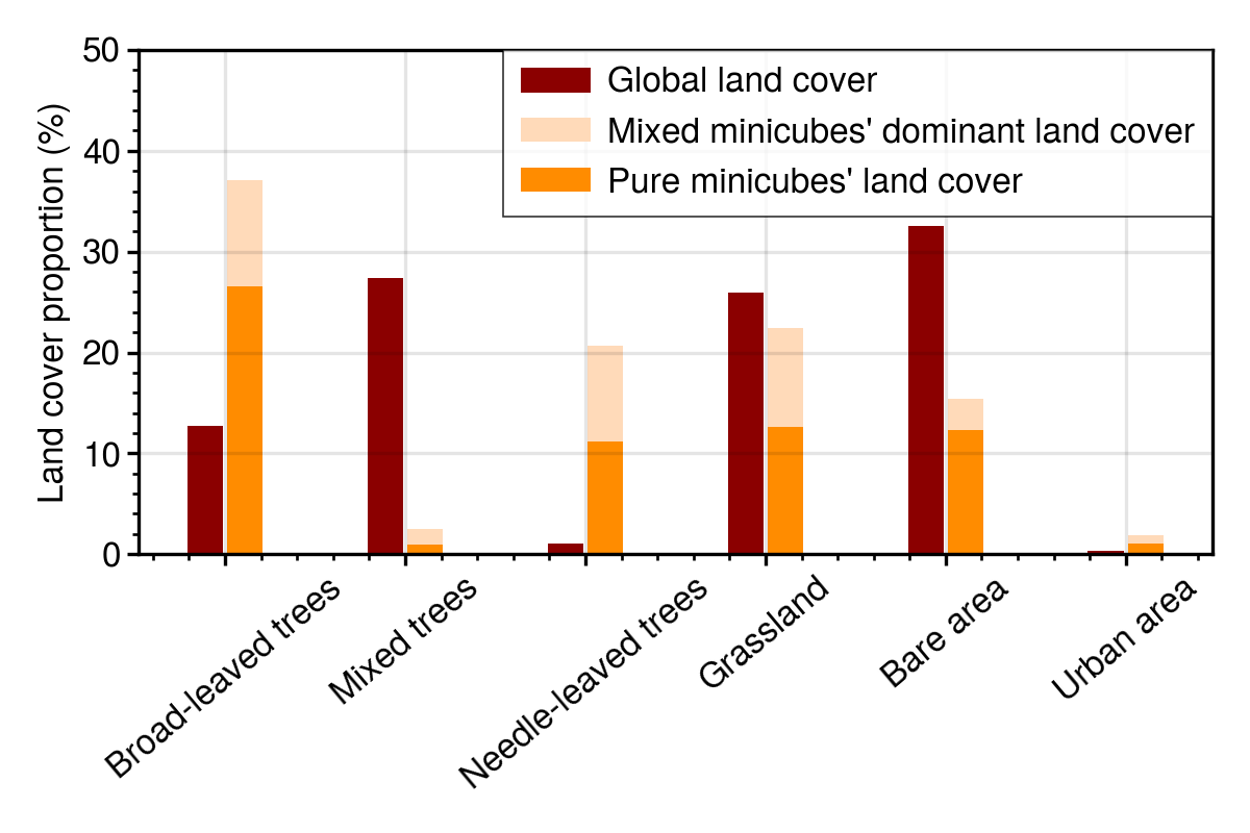}
 \caption{Land cover representation in the \emph{DeepExtremeCubes} dataset. The two plotted datasets use different reference points. The red bars consider the global areas of selected land cover as 100\%, while the light and heavy orange bars treat the total sum of land cover types across all minicubes as 100\%.}
 \label{fig4}
\end{figure}

To validate the land cover representation in the \emph{DeepExtremeCubes} dataset, we compared it with those of the global land cover (see Fig.~\ref{fig4}). Our analysis reveals an overrepresentation of broad-leaved and needle-leaved trees in minicubes compared to the global distribution. Conversely, grassland and urban area align proportionally with the global distribution, while mixed trees samples are underrepresented. This underrepresentation is the result of our method, which defines the dominant land cover in each minicube as its primary land cover, and mixed trees often coincide with either broad-leaved or needle-leaved trees. From the ML perspective, having a larger number of minicubes covered by broad-leaved or needle-leaved trees, as opposed to mixed trees, enhances prediction accuracy.

\subsection*{Spatial analysis of minicube distribution}

To assess the impact of our stratified sampling on the spatial distribution of minicube locations, we computed the spatial autocorrelation of their extreme event occurrences. The analysis yielded a Moran's I value of 0.89, with a p-value of 0.001 and a z-score of 300.48. The high Moran's I value (0.89) reveals a significant and strong positive spatial autocorrelation among minicube locations. This suggests that locations with similar event statuses ("extreme" or "non-extreme") are clustered together. The statistical significance of this clustering is confirmed by the p-value of 0.001 and the high z-score (300.48), indicating that the observed spatial pattern is highly unlikely to be the result of random chance. Additionally, the proximity of "extreme" minicubes to "non-extreme" minicubes demonstrates a correlation in our dataset due to the sampling strategy. This effective sampling approach ensures that "non-extreme" samples are geographically close to "extreme" samples, confirming that our dataset's spatial correlation is preserved.

We analysed the shortest distance along which an "extreme" minicube can find a "non-extreme" minicube with the same land cover and computed the proportion of these minicubes in the total "extreme" minicubes along that distance. It maintains comparable environmental and vegetation characteristics while differing only in the impact of the CHD event on the "extreme" minicube. The results indicate that at a surrounding distance of 200 km, approximately 25\% to 35\% of extreme minicubes of each land cover type (excluding urban areas) can find a non-extreme minicube with the same land cover as a reference. Urban areas have a 10\% probability at this distance, which is attributed to their low coverage in both the global map and minicube coverage (see Fig.~\ref{fig4}). At a surrounding distance of 100 km, extreme minicubes covered by bare areas have a relatively high probability of finding a non-extreme minicube, as bare areas dominate global land coverage. A similar pattern applies to mixed trees. Although the likelihood of finding a paired minicube covered with mixed trees within a surrounding distance of 75 km is relatively low, the high global coverage increases the chances of finding a paired non-extreme minicube at larger distances. For the three main vegetation land covers (broad-leaved trees, needle-leaved trees, and grassland), there is a 5\% to 10\% chance of finding a paired non-extreme minicube with the same land cover type within a surrounding distance of 100 km. Needle-leaved trees slightly exceed grassland at a distance of 150 km, but overall, they exhibit a similar proportion.

\begin{figure}[ht]
 \centering
 \includegraphics[width=0.6\linewidth]{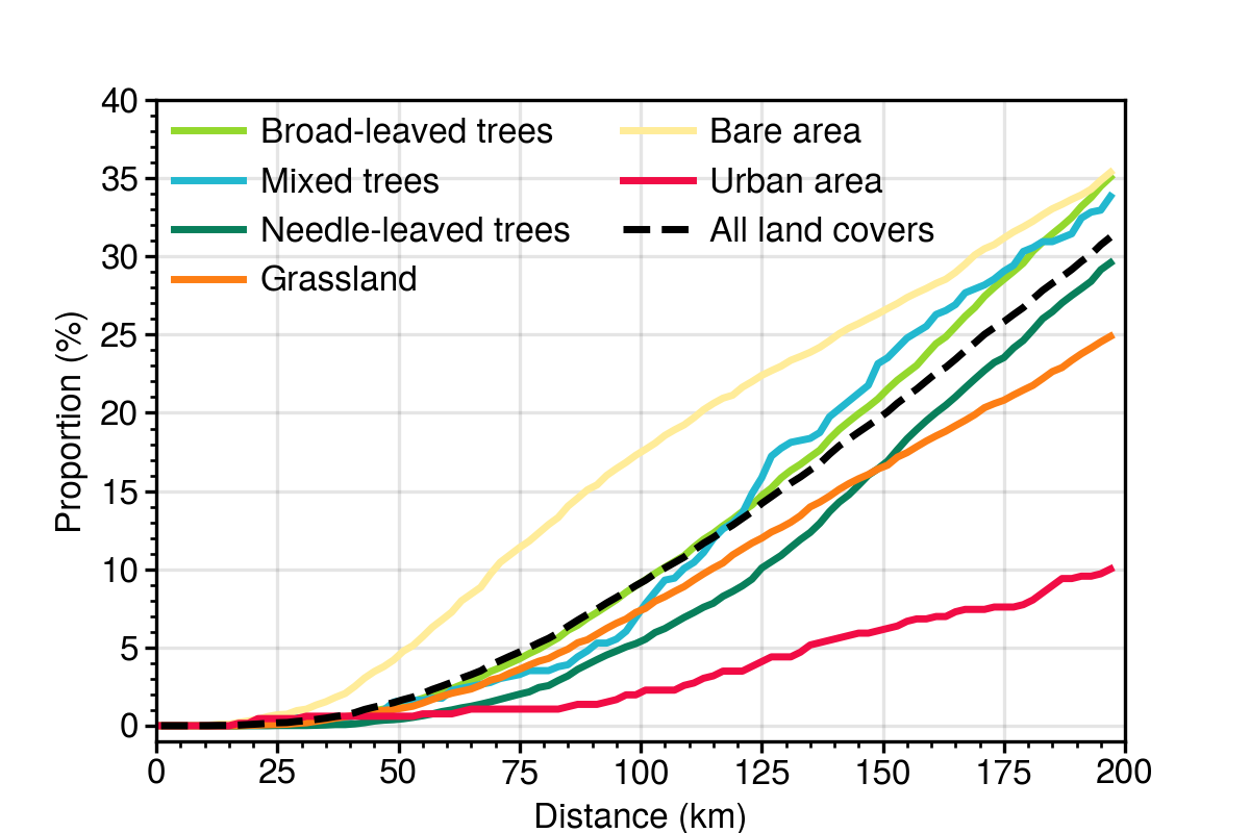}
 \caption{The proportion within a certain distance where an "extreme" minicube can find a "non-extreme" minicube covered by the same land cover.}
 \label{fig5}
\end{figure}

\subsection*{Limitation of the data}
The \emph{DeepExtremeCubes} dataset focuses on forecasting and analysing the impact of CHD extremes on persistent vegetation types. Although we included bare area and urban area as additional land covers, the diversity of land cover in the real world is not fully encompassed. For example, land covers primarily impacted by anthropological effects were omitted (e.g. croplands). This is the trade-off we must make to narrow variables for better prediction and training results in ML models.

The created event days map includes areas experiencing 10 or more event days in the Dheed dataset. This might not be the best approach for a base map for minicube sampling. As mentioned by Weynants et al. (2024)\cite{weynants_dheed_2024}, the amount and volume of extreme events generally follow a power-law distribution, with a few extremely large events and many small ones. Additionally, events in the Dheed dataset with small spatial coverage and short duration could be false alarms. Therefore, selecting the event days map masking areas of more than 10 event days might be simplistic and effective as it accurately reflects the real CHD detection results. Considering this limitation, we propose an alternative strategy, setting criteria that require a volume greater than 1000 units, an area exceeding 30 pixels (equivalent to a 0.25\textdegree increment), and a duration longer than five days. Using these conditions, we identified 114 significant events out of 26,935 events. The resulting event days map is shown in Fig.~s2 in the supplementary information. This alternative map can directly replace the 10-day event map.

\section*{Usage Notes}
\begin{figure}[ht]
 \centering
 \includegraphics[width=0.8\linewidth]{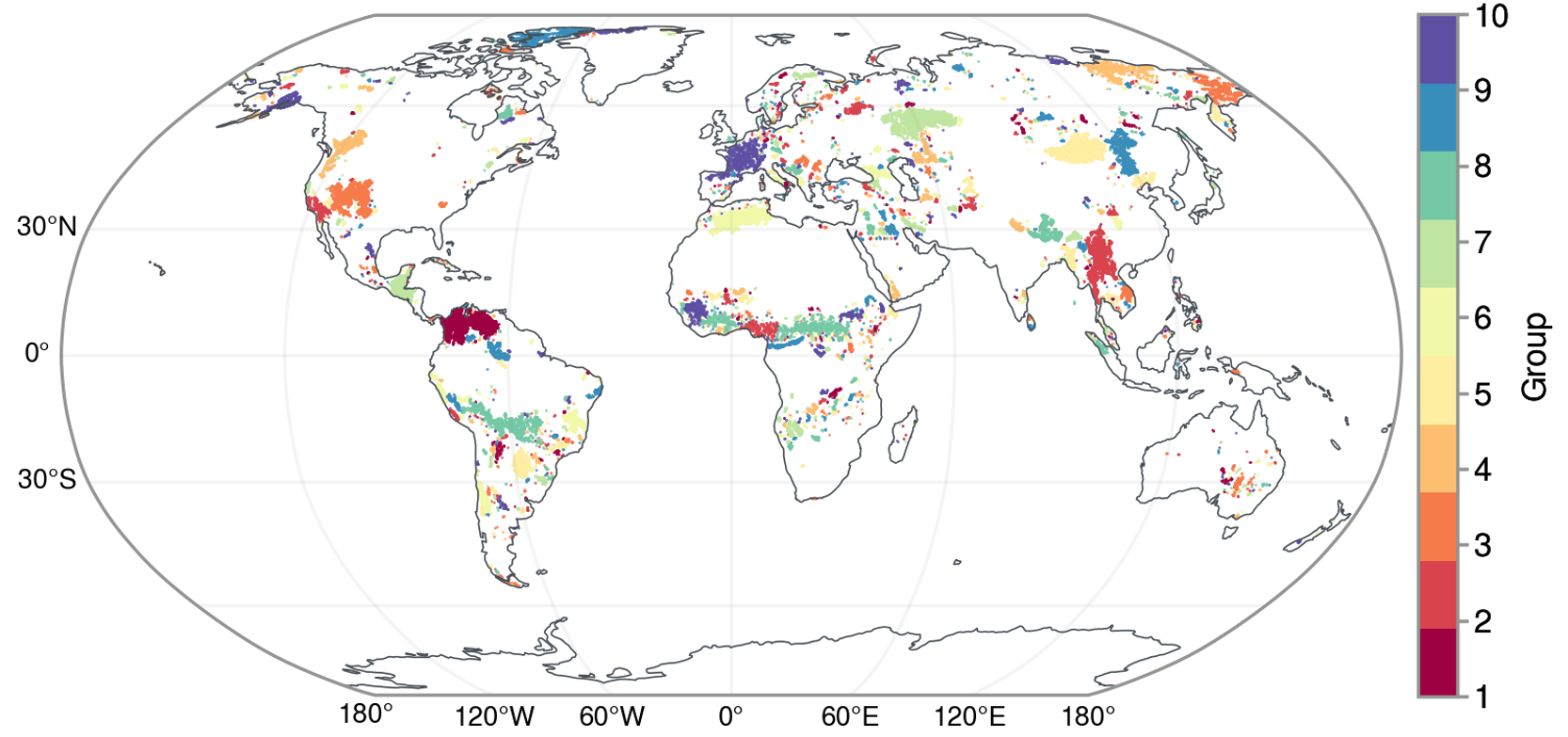}
 \caption{The spatial division of minicube locations into ten folds using a spatially blocked design to reduce autocorrelation between folds.}
 \label{fig:folds}
\end{figure}

The spatial distribution of the minicubes is strongly uneven. Climatic and meteorological data are correlated across large spatial ranges. This leads to minicubes being clustered in and around extreme events. When designing AI methods, one should circumvent this spatial autocorrelation, and samples may not be randomly selected from the minicubes collection to create the training and test sets. To limit the dependence between training, validation and test sets, we implemented a split of the collection into ten folds, ensuring that the distance between minicubes locations from different folds is larger than 50 km\cite{Weynants_DeepExtremes_cv-groups-minicubes_2023}. This split is created in three steps. First, we build a balltree from the haversine distance for all locations (lon, lat). A balltree (also called metric tree) is a tree that is created from successively splitting points into surrounding hyperspheres whose radii are determined from the given metric\cite{omohundro1989balltree,kristoffer_carlsson_kristoffercnearestneighborsjl_2023}. The Haversine distance is the angular distance between two points on the surface of a sphere. Second, we create clusters of locations, ensuring that the distance between locations from different clusters is always larger than 50 km. Third, clusters are distributed into ten groups, in decreasing order of size, always adding the largest cluster to the smallest group. Fig.~\ref{fig:folds} shows the spatial distribution of the resulting groups.

\section*{Code availability}

The code to create the minicubes is hosted at https://github.com/DeepExtremes/minicube-generation.

\bibliography{main.bib}

\section*{Acknowledgements}
The authors acknowledge the support from the ESA AI4Science project
"Multi-Hazards, Compounds and Cascade events: DeepExtremes," 2022-2024, and the European Union's Horizon 2020 research and innovation program within the project "XAIDA: Extreme Events - Artificial Intelligence for Detection and Attribution", (grant agreement 101003469). K.M. acknowledges funding by the Saxon State Ministry for Science, Culture and Tourism (SMWK) – [3-7304/35/6-2021/48880].

\section*{Author contributions statement}

C.J. contributed to conceptualization and methodology, conducted the experiment, software, validation, data curation, writing the original draft, and project administration. T. F. conducted the creation and storage of minicubes. M.W. and F.G. provided expertise in using the Dheed dataset and ran the spatially blocked dataset split. V.B.: conceptualization, methodology. M.A.F.T.: conceptualization, methodology. K.M.: conceptualization, methodology, contributed critically to the drafts, and gave final approval for publication. F.M.: validation. D.M.: software, visualization, writing - review and editing. O.P.: methodology, validation. C.R. contributed to the spatially blocked dataset split. M.M.: conceptualization, supervision, and funding acquisition. All authors contributed to writing - review, and editing.

Authors from the third to the second-to-last are ordered alphabetically.

\end{document}